\documentclass{article}
\usepackage[disable]{todonotes}
\usepackage{graphicx} 
\usepackage{subfigure}
\usepackage{amssymb}

\usepackage{natbib}
\usepackage{multibib}
\usepackage{algorithm}
\usepackage{algorithmic}
\usepackage{amsmath}
\usepackage{hyperref}

\usepackage[accepted]{icml2013} 

\icmltitlerunning{Knapsack Constrained Contextual Submodular List Prediction }

\usepackage{amsthm}
\newtheorem{lemma}{Lemma}
\newtheorem{corollary}{Corollary}

\newcommand{\thmSCP}{2}
\newcommand{\corWM}{2} 
\newcommand{\corSCP}{3}
\newcommand{\secBackground}{3 }
\newcommand{\secContextual}{5 }
\newcommand{\thmSCPKnapsack}{1}
\newcommand{\corSCPKnapsack}{1}

\newtheorem*{thmSCPKnapsackDef}{Theorem \thmSCPKnapsack}
\newtheorem*{corSCPDefKnapsack}{Corollary \corSCPKnapsack}
\DeclareMathOperator*{\E}{\mathbb{E}}
\newenvironment{itemize*}
  {\begin{itemize}
   \setlength{\topsep}{0pt}
    \setlength{\itemsep}{0pt}
    \setlength{\parskip}{0pt}}
  {\end{itemize}}

\begin{document}

\twocolumn[
\icmltitle{Knapsack Constrained Contextual Submodular List Prediction with Application to Multi-document Summarization}

\icmlauthor{Jiaji Zhou}{jiajiz@andrew.cmu.edu}
\icmlauthor{Stephane Ross}{stephaneross@cmu.edu}
\icmlauthor{Yisong Yue}{yisongyue@cmu.edu}
\icmlauthor{Debadeepta Dey}{debadeep@cs.cmu.edu}
\icmlauthor{J. Andrew Bagnell}{dbagnell@ri.cmu.edu}
\icmladdress{School of Computer Science, Carnegie Mellon University, Pittsburgh, PA, USA}

\icmlkeywords{List Optimization, Submodularity, Online Learning}

\vskip 0.3in
]
\begin{abstract}
We study the problem of predicting a \emph{set} or \emph{list} of options under knapsack constraint. The quality of such lists are evaluated by a submodular reward function that measures both quality and diversity. Similar to DAgger \cite{ross2010no}, by a reduction to online learning,  we show how to adapt two sequence prediction models to imitate greedy maximization under knapsack constraint problems: CONSEQOPT \cite{Dey12ConSeqOptArxive} and SCP \cite{Ross2013scp}. Experiments on extractive multi-document summarization show that our approach outperforms existing state-of-the-art methods.

\end{abstract}





\section{Introduction}
Many problem domains, ranging from web applications such as ad placement and content recommendation \cite{yue2011linear},
to identifying successful robotic grasp trajectories \cite{Dey12ConSeqOptArxive}, to extractive multi-document summarization \cite{lin2010multi}, 
require predicting lists of items. Such applications are often budget-limited
and the goal is to choose the best list of items (from a large set of items) with maximal utility. 

In all of these problems, the predicted list should be both relevant
and diverse. For example, in extractive multi-document summarization, one must extract a small set of sentences (as a summary) to match human expert annotations (as measured via  ROUGE \cite{lin2004rouge} statistics). In this setting, selecting redundant sentences will not increase information coverage (and thus the ROUGE score). 
This notion of diminishing returns due to redundancy is often captured formally using submodularity \cite{guestrin08submodtut}. 

Submodular function optimization is intractible. Fortunately, for monotone submodular function, simple forward greedy selection is known to have strong near-optimal performance guarantees and typically works very well in practice \cite{guestrin08submodtut}.  Given access to the monotone submodular reward function, one could simply employ greedy to construct good lists.  

However, in many settings such as document summarization, the reward function is only directly measurable on a finite training set (e.g., where we have expert annotations for computing the ROUGE score).  As such it is increasingly common to take a supervised learning approach, where the goal is to learn a model or policy (based on the training set) that can make good predictions on new test examples where the reward function is not directly measurable.


Prior (state-of-the-art) work on document summarization \cite{lin2010multi,kulesza2012learning}  first learn a surrogate submodular utility function that approximates the ROUGE score, and then perform approximate inference such as greedy using this surrogate function.  While effective, such approaches are only indirectly learning to optimize the ROUGE score.  For instance, small differences between the surrogate function and the ROUGE score may lead to the greedy algorithm performing very differently.

In contrast to prior work, we aim to directly learn to make good greedy predictions, i.e., by learning (on the training set) to mimic the clairvoyant greedy policy with direct access to the reward function. We consider two learning reduction approaches.  Both approaches decompose the joint learning task into a sequence of simpler learning tasks that mimic each iteration of the clairvoyant greedy forward selection strategy.  

The first learning reduction approach decomposes the joint learning a set or list of predictions into a sequence of separate learning tasks \cite{streeter2007online,Radlinski2008ranked,streeter2009online}.
In \cite{deyConSeqOptRSS}, this strategy was extended to the contextual setting
by a reduction to cost-sensitive classification.\footnote{Essentially, each learning problem aims to build a policy that best predicts an item for the corresponding position in the list so as to maximize the expected marginal utility.} 
In the second approach, \cite{Ross2013scp} proposed learning one single policy that applies to each position in the list.  
Both approaches learn to maximize a submodular reward function under simple cardinality constraints, which is unsuitable for settings where different items exhibit different costs.\footnote{In the document summarization setting, different sentences have different lengths.} 

In this paper, we extend both learning reduction approaches to knapsack constrained problems and provide algorithms with theoretical guarantees.\footnote{This is similar to the DAgger approach \cite{Ross11,Ross11b,Ross12} developed for sequential prediction problems like imitation learning and structured prediction. Our approach can be seen as a specialization of this technique for submodular list optimization, and ensures that we learn policies that pick good items under the distribution of list they construct. However, unlike prior work, our analysis leverages submodularity and leads to several modifications of that approach and improved guarantees with respect to the globally optimal list.} 
Empirical experiments on extractive document summarization show that our approach outperforms existing state-of-the-art methods.

\section{Background}

Let $S = \{s_1,\ldots,s_N\}$ denote a set of items, where each item $s_i$ has length $\ell(s_i)$. 
Let $L_1 \subseteq S$ and $L_2 \subseteq S$ denote two sets or lists
of items from $S$.\footnote{Note that we refer to ``lists'' and
  ``sets'' interchangeably in this section.  We often use ``list'' to
  convey an implicit notion of ordering (e.g., the order under which
  our model greedily chooses the list), but the reward function is
  computed over unordered sets.}  Let $\oplus$ denote the list
concatenation operation. We denote $b(s\mid L) = \frac{f(L\oplus s) - f(L)}{\ell(s)}$ as the unit/normalized marginal benefit of adding $s$ to list $L$.

We consider set-based reward functions $f : 2^{|S|} \rightarrow \mathbb{R}^+$ that obeys the following two properties: 
\begin{enumerate}
\item \textbf{Submodularity:} for any two lists $L_1  $, $L_2$ and any item $s$, $f(L_1 \oplus s) - f(L_1) \geq f(L_1\oplus L_2 \oplus s) - f(L_1 \oplus L_2)$.
\item \textbf{Monotonicity:} for any two lists $L_1$, $L_2$, $f(L_1) \leq f(L_1\oplus L_2)$ and  $f(L_2) \leq f(L_1\oplus L_2)$. 
\end{enumerate}
Intuitively, submodularity corresponds to a diminishing returns property and monotinicity indicates that adding more items never reduces the reward. We assume for simplicity that $f$ takes values in [0,1], and in particular $f(\emptyset ) =0$. 

We further enforce a knapsack constraint, i.e., that the computed list $L$ must obey 
$$\ell(L) = \sum_{s \in L} \ell(s) < W,$$
where $W$ is a pre-specified budget.

For the multi-document summarization application,
$S$ refers to the set of all sentences in a summarization task, and $\ell(s)$ refers to the byte length of sentence $s$. The reward function $f$ is then the ROUGE Unigram Recall score, which can be easily shown to be monotone submodular \cite{lin2011class}.

\section{Contextual Submodular Sequence Prediction}

We assume to be given a collection of states $x_1,\ldots,x_T$, where each $x_t$ is sampled i.i.d. from a common (unknown) distribution $D$.  Each state $x_t$ corresponds to a problem instance (e.g., a document summarization task) and is associated with observable features or context.  
We further assume that features describing partially contructed lists are also observable.

We consider learning a sequence of $k$ different policies $ L_{\pi,k} = (\pi_1,
\pi_2, ... ,\pi_k )$ with the goal of applying them sequentially to
predict a list $L_t$ for $x_t$:   policy $\pi_i$ takes as input the
features of $x_t$ and outputs an item $s$ in $S_t$ to append as the
$i$th element after $L_{t,i-1}$. Therefore $L_{\pi,k}$ will produce a
list $L_t = \{ \pi_1(x_t) , \pi_2(x_t) ,..., \pi_k(x_t) \}$. We also
consider the case that a single policy $\pi$ takes input of both the
features of $x_t$ and $L_{t,i-1}$ such that it produces a list  $L_t =
\{ \pi(x_t, L_{t,0}) , \pi(x_t, L_{t,1}) ,..., \pi(x_t, L_{t,k-1}) \}$.
We refer to $\pi^t$ as the online learner's current policy when predicting for state $x_t$ (which can be either a list of policies or a single policy depending on the algorithm).  For both cases (described below), we show how to extend them to deal with knapsack constraints.


\begin{algorithm}[t]
\begin{small}
\begin{algorithmic}
\STATE \textbf{Input:} policy class $\Pi $, budget length $W$.
\STATE Pick initial policy $\pi_1$ 
\FOR{$t=1$ \textbf{to} $T$}
\STATE  Observe features of a sampled state $x_t \sim D$ and item set $S_t$
\STATE  Construct list $L_t$  using $\pi_t$.
\STATE  Define $|L_t|$ new cost-sensitive classification examples $\{(v_{ti},c_{ti},w_{ti})\}_{i=1}^{|L_t|}$ where:  
\begin{itemize}
\item $v_{ti}$ is the feature vector of state $x_{t}$ and list $L_{t,i-1}$
\item $c_{ti}$ is a cost vector such that $\forall s \in S_t:\ c_{ti}(s) = \max_{s' \in \mathcal{S_t}} b(s'|L_{t,i-1},x_t) - b(s|L_{t,i-1},x_t)  $
\item $w_i =  [\prod_{j=i+1}^{|L_t|}(1 - \frac{\ell(s_{t,j})}{W} )] \ell(s_{t,i})$ is the weight of this example 
\end{itemize}
\STATE $\pi^{t+1} = \textsc{Update}(\pi^t, \{(v_{ti},c_{ti},w_{ti})\}_{i=1}^{|L_t|})$
\ENDFOR
\STATE \textbf{return} $\pi_{T+1}$
\end{algorithmic}
\end{small}
\caption{Knapsack Constrained Submodular Contextual Policy Algorithm. \label{algSCPKnapsack}}
\end{algorithm}

\subsection{CONSEQOPT: Learning a sequence of (different) policies}
  CONSEQOPT~\cite {Dey12ConSeqOptArxive}  learns a sequence of policies under cardinality constraint by reducing the learning problem to $k$ separate supervised cost-sensitive classification problems in batch. 
We consider the knapsack constraint case and provide and error bound derived from regret bounds for online training. 

\subsection{SCP: Learning one single policy}
  SCP \cite {Ross2013scp} learns one single policy that applies to each position for the list. The algorithm and theoretical analysis apply to cardinality constrained problems. Under the same online learning reduction framework \cite{ross2010no}, we extend the analysis and algorithm to the knapsack constraint setting.    

\subsection{Algorithm}\label{secAlgo}

Algorithm \ref{algSCPKnapsack} shows our algorithm for both knapsack constrained CONSEQOPT and SCP.  
At each iteration, SCP/CONSEQOPT constructs a list $L_t$ for state $x_t$ using its current policy/list of policies. We get the observed benefit of each item in $S_t$ at every position of the list $L_t$, organized as $|L_t|$ sets of cost sensitive classification examples $\{(v_{ti},c_{ti},w_{ti})\}_{i=1}^{|L_t|}$, each consisting of $|S_t|$ instances. These new examples are then used to update the policy. 
Note that the online learner's update operation (\textsc{Update}) is different for CONSEQOPT and SCP. CONSEQOPT has one online learner for each of its position-dependent policy and  $\{(v_{ti},c_{ti},w_{ti})\}$ is used to update the $i$th online learner, while SCP would use all of $\{(v_{ti},c_{ti},w_{ti})\}_{i=1}^{|L_t|}$ to update a single online learner.

\subsubsection{Reduction to Ranking}
In the case of a finite policy class $\Pi$, one may leverage algorithms like Randomized Weighted Majority and update the distribution of policies in $\Pi$.  However, achieving no-regret for a policy class that has infinite number of elements is generally intractable. 
As mentioned above, both learning reduction approaches reduce the problem to a better-studied learning problem, such as cost-sensitive classification.\footnote{The reduction is implemented via the \textsc{Update} subroutine in Algorithm \ref{algSCPKnapsack}.}

We use a reduction to ranking that penalizes ranking an item $s$ above another better item $s'$ by an amount proportional to their difference in cost. Essentially, for each cost-sensitive example $(v,c,w)$, we generate $\mathcal{|S|}(\mathcal{|S|} -1)/2$ ranking examples, one for every distinct pair of items $(s,s')$, where we must predict the best item among $(s, s')$ (potentially by a margin), with a weight of $w|c(s)-c(s')|$. 

For example, if we train a linear SVM with feature weight $h$, this would be translated into a weighted hinge loss of the form: $w |c(s)-c(s')| \max( 0, 1 - h^\top (v(s)-v(s')) \textrm{sign}(c(s)-c(s')) )$. At prediction time we simply predict the item $s^*$ with highest score, $s^* = \arg\max_{s \in \mathcal{S}} h^\top v(s)$. 

This reduction is equivalent to the Weighted All Pairs reduction \cite{beygelzimer2005error} except we directly transform the weighted 0-1 loss into a convex weighted hinge-loss upper bound -- this is known as using a convex \textit{surrogate} loss function.   This reduction is often advantageous whenever it is easier to learn relative orderings rather than precise cost.

\subsection{Theoretical Guarantees}

We now present theoretical guarantees of Algorithm 1 relative to a randomization of an optimal policy $\tilde{L}_{\pi}^*$ that takes the following form. Let $L_{\pi}^* = (\pi^*_1,\ldots,\pi^*_W)$ denote an optimal deterministic policy list of size $W$. 
Let $\tilde{L}_{\pi}^*$ denote a randomization of $L_{\pi}^*$ that generates predictions $L_t$ in the following way:  
Apply each $\pi_i^* \in L_{\pi}^*$ sequentially to $x_t$, and include the prediction $\pi^*_i(x_t)$ picked by $\pi^*_i$ with probability probability $p = {1}/{\ell(\pi^*_i(x_t))}$, or otherwise discard.  Thus, we have

\begin{displaymath}
L_{t,i} = \left\{ 
  \begin{array}{ll}
    L_{t,i-1} \oplus \pi^*_i(x_t)  &  \mbox{w.p. } \frac{1}{\ell(\pi^*_i(x_t))} \\[5pt]
    L_{t,i-1}  & \mbox{w.p. } 1 - \frac{1}{\ell(\pi^*_i(x_t))}
  \end{array} \right.
\end{displaymath}

We can also think of each policy as having probability of being executed to be inversely proportional to the cost of the element it picks. Therefore, in expectation, each policy will add the corresponding normalized/unit benefit to the reward function value. 

Ideally, we would like to prove theoretical guarantees relative to the actual deterministic optimal policy. However,  $\tilde{L}_{\pi}^*$ can be intuitively seen as an average behavior of deterministic optimal policy. We defer an analysis comparing our approach to the deterministic optimal policy to future work.

We present learning reduction guarantees that relate performance on our actual submodular list optimization task to the regret of the corresponding online cost-sensitive classification task.
Let  $\{ \ell_t  \}_{t=1}^T$ denote a sequence of losses in the corresponding online learning problem, where $\ell_t : \Pi \rightarrow \mathbb{R^+}$ represents the loss of each policy $\pi$ on the cost-sensitive classification examples $\{v_{ti},c_{ti},w_{ti}\}_{i=1}^|L_t|$ collected in Algorithm \ref{algSCPKnapsack} for state $x_t$. The accumulated regret incurred by the online learning subroutine (\textsc{Update}) is denoted by $R = \sum_{t=1}^T \ell_t(\pi^t) - \min_{\pi \in \Pi} \sum_{t=1}^T \ell_t(\pi)$. 

Let $F(\pi) =  \mathbb{E}_{x \sim D} [ f_x(\pi(x)) ]$ denote the expected value of the lists constructed by $\pi$. Let $\hat{\pi} = \arg\max_{t \in \{1,2,\dots,T\}} F(\pi^t)$ be the best policy found by the algorithm, and define the mixture distribution $\overline{\pi}$ over policies such that $F(\overline{\pi}) = \frac{1}{T} \sum_{t=1}^T F(\pi^t)$.  

We will focus on showing good guarantees for $F(\overline{\pi})$, as $F(\hat{\pi}) \geq F(\overline{\pi})$. We now show that, in expectation, $\overline{\pi}$ (and thus $\hat{\pi}$) must construct lists with performance guarantees close to that of the greedy algorithm over policies in $\Pi$ if a no-regret subroutine is used: 

\begin{thmSCPKnapsackDef}
  After $T$ iterations, for any $\delta \in (0,1)$, we have that with probability at least $1-\delta$:
   \begin{displaymath}
   F(\overline{\pi}) \geq (1-1/e)F(\tilde{L}_{\pi}^*) - \frac{R}{T} - 2 \sqrt{\frac{2 \ln(1/\delta)}{T}}
   \end{displaymath}
\end{thmSCPKnapsackDef}
Theorem 1 implies that the difference in reward  between our learned
policy and the (randomization of the) optimal policy is upper bounded
by the regret of the online learning algorithm used in \textsc{Update}
divided by $T$, and a second term that shrinks as $T$ grows.
\footnote{Note that the list that policy $\pi$ produces may not always
just fit into budget $W$ since the last element it chooses may exceed
budget. In practice, we always choose the element that is most
favoured by the policy according to the its ranking over items that
still fit into the budget. One can show that this additional step has
the same $(1 - 1/e)$ approximation guarantee with respect to the
optimal stochastic list of policies of size $W/2$. This is because we can eliminate all
items that have cost greater than $W/2$ as a pre-processing. }

Running any no-regret online learning algorithm such as Randomized Weighted Majority \cite{littlestone1994weighted} in \textsc{Update} would guarantee 
$$\frac{R}{T} = O\left(\frac{\sqrt{WgT\ln{|\Pi|}} }{T}\right)$$ for SCP and CONSEQOPT,\footnote{Naively, CONSEQOPT would have average regret that scales as $O\left(\frac{k(\sqrt{WgT\ln{|\Pi|}})}{T}\right)$, since we must run $k$ separate online learners. However, similar to lemma 4 in \cite{streeter2007online} and corollary 2 in \cite{Ross2013scp}, it can be shown that the average regret is the same as SCP.} where $g$ is the largest possible normalized/unit marginal benefit. 
Note that when applying this approach in a batch supervised learning setting, $T$ also corresponds to the number of training examples.

\subsubsection{Convex Surrogate Loss Functions}
Note that we could also use an online algorithm that uses surrogate convex loss functions (e.g., ranking loss) for computational efficiency reasons when dealing with infinite policy classes. 
As in \cite{Ross2013scp}, we provide a general theoretical result that applies if the online algorithm is used on any convex upper bound of the cost-sensitive loss. An extra penalty term is introduced that relates the gap between the convex upper bound to the original cost-sensitive loss:

\begin{corSCPDefKnapsack}
If we run an online learning algorithm on the sequence of convex losses $C_t$ instead of $\ell_t$, then after $T$ iterations, for any $\delta \in (0,1)$, we have that with probability at least $1-\delta$:
\begin{displaymath}
F(\overline{\pi}) \geq (1-1/e) F(\tilde{L}_{\pi}^*) - \frac{\tilde{R}}{T} - 2 \sqrt{\frac{2 \ln(1/\delta)}{T}} - \mathcal{G}
\end{displaymath}
where $\tilde{R}$ is the regret on the sequence of convex losses $C_t$, and $\mathcal{G} = \frac{1}{T}[\sum_{t=1}^T (\ell_t(\pi^t) - C_t(\pi^t)) + \min_{\pi \in \tilde{\Pi}} \sum_{t=1}^T C_t(\pi) - \min_{\pi' \in \tilde{\Pi}} \sum_{t=1}^T \ell_t(\pi')]$ is the  ``convex optimization gap'' that measures how close the surrogate losses $C_t$ are to minimizing the cost-sensitive losses $\ell_t$.
\end{corSCPDefKnapsack}
	
  The gap $\mathcal{G}$ may often be small or non-existent. For instance, in the case of the reduction to regression or ranking, $\mathcal{G} = 0$ in realizable settings where there exists a predictor which models accurately all the costs or accurately ranks the items by a margin. Similarly, in cases where the problem is near-realizable we would expect $\mathcal{G}$ to be small. We emphasize that this convex optimization gap term is not specific to our particular scenario, but is (implicitly) always present whenever one attempts to optimize classification accuracy, e.g. the 0-1 loss, via convex optimization.\footnote{For instance, when training a SVM in standard batch supervised learning, we would only expect that minimizing the hinge loss is a good surrogate for minimizing the 0-1 loss when the analogous convex optimization gap is small.} This result implies that whenever we use a good surrogate convex loss, then using a no-regret algorithm on this convex loss will lead to a policy that has a good approximation ratio to the optimal list of policies.


\section{Application to Document Summarization}

We apply our knapsack versions of the SCP and CONSEQOPT algorithms to an extractive multi-document summarization task. Here we construct summaries subject to a maximum budget of characters $W$ by 
extracting sentences in the same order of occurrence as in the original document.

Following the experimental set up from previous work of \cite{lin2010multi} (which we call SubMod) and \cite{kulesza2012learning} (which we call DPP), we use the datasets from the Document
Understanding Conference (DUC) 2003 and 2004 (Task 2) \cite{dang2005overview}. The data consists of clusters of documents, where each cluster contains
approximately $10$ documents belonging to the same topic and four human reference summaries. We train on the 2003 data (30 clusters) and test on the 2004 data (50 clusters). The budget length is 665 bytes, including spaces.

We use the ROUGE \cite{lin2004rouge} unigram statistics (ROUGE-1R, ROUGE-1P,
ROUGE-1F) for performance evaluation. Our method directly attempts learn a policy that optimizes the
ROUGE-1R objective with respect to the reference summaries, which can be easily
shown to be monotone submodular \cite{lin2011class}.



Intuitively, we want to predict sentences that are both short and capture a diverse set of important concepts in the target summaries. This is captured in our definition of cost using the difference of normalized benefit $c_{ti}(s) = b(s|L_{t,i-1},x_t) - \max_{s' \in \mathcal{S}} b(s'|L_{t,i-1},x_t)$. We use a reduction to ranking as described in Section \ref{secAlgo}.\footnote{We use Vowpal Wabbit \cite{vw} for online training and the parameters for online gradient descent are set as default. } 

\subsection{Feature Representation}

The features for each state/document $x_t$ are sentence-level features. 
Following \cite{kulesza2012learning}, we consider features $f_i$ for each sentence
consisting of \emph{quality features} $q_i$ and \emph{similarity features} $\phi_i$ ($f_i=[q_i^T,\phi_i^T]^T$). The quality features, attempt to capture the
representativeness for a single sentence. We use the same quality features as in \cite{kulesza2012learning}. 

Similarity features $q_i$ for sentence $s_i$ as we construct the list $L_t$ measure a notion of distance of a proposed sentence to sentences already included in the list.
A variety of similarity features were considered, the simplest being average squared distance of tf-idf vectors. Performance varied little depending on the details of these features. 
The experiments presented use three types: 1) following the idea in \cite{kulesza2012learning} of similarity
as a volume metric, we compute the
squared volume of the parallelopiped spanned by the TF-IDF vectors of
sentences in the set $L_{t,k} \cup {s_i}$, which is equivalent to the determinant of submatrix $\mbox{det}(G_{L_{t,k} \cup {s_i}})$ of the Gram Matrix $G$, whose elements are pairwise TF-IDF vector inner products; 
2) the product between $\mbox{det}(G_{L_{t,k} \cup {s_i}})$ and the quality features; 3) the minimum absolute distance of quality features between $s_i$ and each of the elements in $L_{t,k}$.


\subsection{Results}
  
Table \ref{DocSumTable} documents the performance (ROUGE unigram statistics) of knapsack constrained SCP and CONSEQOPT compared with SubMod and DPP (which are both state-of-the-art approaches). 
``Greedy (Oracle)'' corresponds to the oracle used to train DPP, CONSEQOPT and SCP. 
This method directly optimizes the test ROUGE score and thus serves as an upper bound on this class of techniques.
We observe that both SCP and CONSEQOPT outperform SubMod and DPP in terms of all three ROUGE Unigram statistics. 



\begin{table}
    \tiny
\centering
  \begin{tabular}{|l || c | c | c | }
    \hline
    \bf{System} & \bf{ROUGE-1F} & \bf{ROUGE-1P} & \bf{ROUGE-1R} \\ 
    \hline 
    SubMod & 37.39 & 36.86 & 37.99 \\ 
    \hline
    DPP  & 38.27 & 37.87 & 38.71 \\
    \hline
    CONSEQOPT & $ 39.02\pm0.07$ & 39.08$\pm 0.07$ & 39.00$\pm0.12$ \\
    \hline
    SCP  & \textbf{39.15$\pm0.15$} & \textbf{39.16$\pm 0.15$} & \textbf{39.17$\pm0.15$} \\
    \hline
    Greedy (Oracle) & 44.92 & 45.14 & 45.24 \\
    \hline
  \end{tabular}
\caption{ROUGE unigram statistics on the DUC 2004 test set}
  \label{DocSumTable}
\end{table}
\normalsize

\subsection*{Acknowledgements}
\vspace{-0.06in}
\begin{small}
This research was supported by NSF NRI \emph{Purposeful Prediction} and the Intel Science and Technology Center on Embedded Computing.  
We gratefully thank
Martial Hebert for valuable discussions and Alex Kulesza for providing data and code. 
\end{small}

\bibliography{references}
\bibliographystyle{icml2013}

\clearpage

\section*{Appendix - Proofs of Theoretical Results}

This appendix contains the proofs of  theoretical results presented in this paper. We also encourage readers to refer to \cite{Ross2013scp} and its supplementary materials for the proof of the cardinality constrained case. 

\subsection*{Preliminaries}

We begin by proving a number of lemmas about monotone submodular functions, which will be useful to prove our main results. Note that we refer to ``lists'' and ``sets'' interchangeably in this section.  We often use ``list'' to convey an implicit notion of ordering (e.g., the order under which our model greedily chooses the list), but the reward function is computed over unordered sets.

\begin{lemma} \label{lemAddList}
Let $\mathcal{S}$ be a set and $f$ be a monotone submodular function defined on a list of items from $\mathcal{S}$. For any lists $A,B$, we have that:
\begin{displaymath}
f(A \oplus B) - f(A) \leq |B| ( \mathbb{E}_{s \sim U(B)}[f(A \oplus s)] - f(A) )
\end{displaymath}
where $U(B)$ denotes the uniform distribution on items in $B$. 
\end{lemma}
\begin{proof}
For any two lists $A$ and $B$, let $B_i$ denote the list of the first $i$ items in $B$, and $b_i$ the $i^{th}$ item in $B$. We have that:
\begin{displaymath}
\begin{array}{rl}
\multicolumn{2}{l}{f(A \oplus B) - f(A)}\\
= & \sum_{i=1}^{|B|} f(A \oplus B_i) -f(A \oplus B_{i-1})\\
\leq & \sum_{i=1}^{|B|} f(A \oplus b_i) -f(A) \\
= & |B| ( \mathbb{E}_{b \sim U(B)}[f(A \oplus b)] - f(A) )\\
\end{array}
\end{displaymath}
where the inequality follows from the submodularity property of $f$.
\end{proof}
 
\begin{corollary} \label{corAddListStochastic}
Let $\mathcal{S}$ be a set and $f$ be a monotone submodular function defined on a list of items from $\mathcal{S}$.  Let $\tilde{B} = \{\tilde{b}_1,\ldots,\tilde{b}_{|B|}\}$ denote a stochastic list generated stochastically from the corresponding deterministic list $B$ as follows:
 $$\forall i: \tilde{b}_i = 
 \left\{\begin{array}{ll}
b_i & \mbox{w.p. } \frac{1}{\ell(b_i)}\\
\o \mbox{ (empty)} & \mbox{otherwise}
\end{array}\right..$$
Then we have that:
\begin{displaymath}
\mathbb{E}[f(A \oplus \tilde{B})] - f(A) \leq |B|  \mathbb{E}_{s \sim U(B)}\left[\frac{f(A \oplus s) - f(A)}{\ell(s)}\right],
\end{displaymath}
where the first expectation is taken over then randomness of $\tilde{B}$, and  $U(B)$ denotes the uniform distribution on items in $B$. 
\end{corollary}

\begin{proof}
\begin{displaymath}
\begin{array}{rl}
\multicolumn{2}{l}{ \mathbb{E} [f(A \oplus \tilde{B})] - f(A)} \\
= & \sum_{i=1}^{|\tilde{B}|} \mathbb{E} [f(A \oplus \tilde{B}_i)] - \mathbb{E} [f(A \oplus \tilde{B}_{i-1})]\\
\leq & \sum_{i=1}^{|\tilde{B}|} \mathbb{E} [f(A \oplus \tilde{b}_i)] -f(A) \\
= &\sum_{i=1}^{|\tilde{B}|}  \frac{1}{\ell(b_i)} f(A \oplus b_i) + (1 - \frac{1}{\ell(b_i)}) f(A) - f(A) \\ 
= &\sum_{i=1}^{|B|}  \frac{f(A \oplus b_i) - f(A)}{\ell(b_i)} \\
= & |B|  \mathbb{E}_{b \sim U(B)}[ \frac{f(A \oplus b) - f(A)}{\ell(b)}  ]\\
\end{array}
\end{displaymath}
where the inequality follows from the submodularity property of $f$.
\end{proof}

\begin{lemma} \label{lemBudgetErrorKnapsack}
Let $\mathcal{S}$, $f$, $A$, $\tilde{B}$, $B$, and $U(B)$ be defined as in Corollary \ref{corAddListStochastic}.
Let $\ell(A) = \sum_{i=1}^{|A|} \ell(a_i)$ denote the sum of length of each element $a_i$ in $A$, and let $A_j$ denote the list of the first $j$ items in $A$. Define $\epsilon_j = \mathbb{E}_{s \sim U(B)}[\frac{f(A_{j-1} \oplus s) - f(A_{j-1})} {\ell(s)}] - \frac{f(A_j) - f(A_{j-1})}{\ell(a_j)}$ as the additive error term in competing with the average marginal normalized benefits of the items in $B$ when picking the $j^{th}$ item in $A$ (which could be positive or negative). Then for $\alpha = \exp{(-\ell(A)/|B|)}$, we have
\begin{displaymath}
f(A) \geq (1 - \alpha) \E [f(\tilde{B})] - \sum_{i=1}^{|A|} \left[\prod_{j=i+1}^{|A|}\left(1 - \frac{\ell(a_j)}{|B|}\right)\right] \ell(a_i)\epsilon_i.
\end{displaymath}
\end{lemma}

\begin{proof}
Using the monotonicity property of $f$ and Corollary \ref{corAddListStochastic}, we have that: 
\begin{align*}
\mathbb{E}[f(\tilde{B})] - f(A) & \leq \mathbb{E}[f(A \oplus \tilde{B})] - f(A)\\
   & \leq |B| \mathbb{E}_{s \sim U(B)}\left[\frac{f(A \oplus s) - f(A)}{\ell(s)}\right].
\end{align*}

Define $\Delta_j = \E[f(\tilde{B})] - f(A_j)$. We have that:
\begin{align*}
\Delta_j & \leq   |B| \mathbb{E}_{s \sim U(B)}\left[\frac{f(A_j \oplus s) - f(A_j)}{\ell(s)}\right] \\
&=  |B| \mathbb{E}_{s \sim U(B)} \left[ \frac{f(A_j \oplus s) - f(A_j)}{\ell(s)}\right.  \\
&\ \ \ \ \ \left.- \frac{f(A_{j+1}) - f(A_j)}{\ell(a_{j+1})}+ \frac{f(A_{j+1}) - f(A_j)}{\ell(a_{j+1})}\right] \\
& = \frac{|B|}{\ell(a_{j+1})} \left(\ell(a_{j+1})\epsilon_{j+1} + \Delta_j - \Delta_{j+1} \right). 
\end{align*}
Rearranging the terms yields
\begin{align*}
\frac{\ell(a_{j+1})}{|B|}\Delta_j & \leq \ell(a_{j+1})\epsilon_{j+1} + \Delta_j - \Delta_{j+1}\\
   \Delta_{j+1} &\leq \left(1 - \frac{\ell(a_{j+1})}{|B|}\right) \Delta_j + \ell(a_{j+1}) \epsilon_{j+1}.
\end{align*}
Recursively expanding, we get
\begin{align*}
&\Delta_{|A|} \leq \\
&   \ \ \ \prod_{i=1}^{|A|} \left(1- \frac{\ell(a_i)}{|B|} \right)\Delta_0 + \sum_{i=1}^{|A|} \prod_{j=i+1}^{|A|} \left(1 - \frac{\ell(a_j)}{|B|}\right)\ell(a_i)\epsilon_i.
   \end{align*}
The term $\prod_{i=1}^{|A|} (1- \frac{\ell(a_i)}{|B|} )$ is maximized when all $\ell(a_i)$ are equal, therefore $ \prod_{i=1}^{|A|} (1- \frac{\ell(a_i)}{|B|} ) \leq (1 - \frac{\ell(A)}{|A||B|})^{|A|} \leq \exp(|A|\log(1 - \frac{\ell(A)}{|A||B|})) \leq \exp(-|A|\frac{\ell(A)}{|A||B|}) = \alpha$. Rearranging the terms and using the definition of $\Delta_{|A|} = f(\tilde{B}) - f(A)$ and $\Delta_{0} = f(\tilde{B})$ prove the statement. 
\end{proof}

\subsection*{Proofs of Main Results}

We now provide the proofs of the main results in this paper. We refer the reader to the notation defined in section \secBackground and \secContextual for the definitions of the various terms used.

\begin{thmSCPKnapsackDef}
  After $T$ iterations, for any $\delta \in (0,1)$, we have that with probability at least $1-\delta$:
   \begin{displaymath}
   F(\overline{\pi}) \geq (1-1/e) F(\tilde{L}^*_{\pi}) - \frac{R}{T} - 2 \sqrt{\frac{2 \ln(1/\delta)}{T}} 
   \end{displaymath}
\end{thmSCPKnapsackDef}
\begin{proof}
\begin{small}
\begin{align*}
F(\bar{\pi}) &	= \frac{1}{T} \sum_{t=1}^T F(\pi^t)\\
&	= \mathbb{E}_{x \sim D}\left[ \frac{1}{T}\sum_{t=1}^T   f_x(\pi^t(x)) \right] \\
&	= \left(1-\frac{1}{e}\right)\mathbb{E}_{x \sim D}[f_x(\tilde{L}^*_{\pi}(x))] \\
&\ \ \ \ 	 - \mathbb{E}_{x \sim D}\left[\left(1-\frac{1}{e}\right)f_x(\tilde{L}^*_{\pi}(x))  - \frac{1}{T}\sum_{t=1}^T f_x(\pi^t(x)) \right]  \\
&	=  \left(1-\frac{1}{e}\right)\mathbb{E}_{x \sim D}[f_x(\tilde{L}^*_{\pi}(x))]  \\
&\ \ \ \ 	 - \left(1-\frac{1}{e}\right)\frac{1}{T} \sum_{t=1}^T \left(f_{x_t}(\tilde{L}^*_{\pi}(x_t)) - f_{x_t}(L_t)\right)\\
&\ \ \ \ 	 - \frac{1}{T}\sum_{t=1}^{T} X_t, 
\end{align*}
\end{small}
	 where 
\begin{align*}
    X_t & = (1-1/e) \{ \mathbb{E}_{x \sim D}[f_x(\tilde{L}^*_{\pi}(x))]\\
        &\ \ \ \ - f_{x_t}(\tilde{L}^*_{\pi}(x_t)) \} - \{\mathbb{E}_{x \sim D} [ f_x(\pi^t(x)) ]  - f_{x_t}(L_t) \}.
  \end{align*}
Because each $x_t$ is sampled i.i.d. from $D$, and the distribution of policies used to construct $L_t$ only depends on $\{ x_\tau \}_{\tau=1}^{t-1}$ and $\{ L_{\tau} \}_{\tau=1}^{t-1}$, then each $X_t$ when conditioned on $\{ X_{\tau} \}_{\tau=1}^{t-1}$ will have expectation 0, and because $f_x \in [0,1]$ for all state $x \in \mathcal{X}$, $X_t$ can vary in a range $r \subseteq [-2,2]$. Thus the sequence of random variables $Y_t = \sum_{i=1}^t X_i$ forms a martingale where $|Y_t - Y_{t+1}| \leq 2$. By the Azuma-Hoeffding's inequality, we have that 
   $$P(Y_T/T \geq \epsilon) \leq \exp(-\epsilon^2 T/8).$$
   Hence for any $\delta \in (0,1)$, we have that with probability at least $1-\delta$,  $Y_T/T = \frac{1}{T}\sum_{t=1}^T{X_t} \leq 2 \sqrt{\frac{2 \ln(1/\delta)}{T}}$. Hence we have that with probability at least $1-\delta$:
	\begin{displaymath}
      	\begin{array}{rl}
      		\multicolumn{2}{l}{F(\overline{\pi})} \\
    		\geq & (1-1/e)\mathbb{E}_{x \sim D}[f_x(\tilde{L}^*_{\pi}(x))]  \\
    			& - [(1-1/e)\frac{1}{T} \sum_{t=1}^T f_{x_t}(\tilde{L}^*_{\pi}(x_t)) \\
    			&  - \frac{1}{T}\sum_{t=1}^T f_{x_t}(L_t) ] - 2 \sqrt{\frac{2 \ln(1/\delta)}{T}}  \\
      		\end{array}	
	\end{displaymath}
	Let $s_{t,i}$ denote the $i$th element in $L_t$ and $w_i =  [\prod_{j=i+1}^{|L_t|}(1 - \frac{\ell(s_{t,j}))}{k} )] \ell(s_{t,i}) $. From lemma \ref{lemBudgetErrorKnapsack}, we have: 
	\begin{displaymath}
	\begin{array}{rl}
	\multicolumn{2}{l}{(1-1/e)\frac{1}{T} \sum_{t=1}^T f_{x_t}(\tilde{L}^*_{\pi}(x_t))  - \frac{1}{T}\sum_{t=1}^T f_{x_t}(L_t)}\\
	\leq & \frac{1}{T}\sum_{t=1}^T\sum_{i=1}^{|L_t|} w_i (\mathbb{E}_{\pi \sim U(L^*_{\pi})}[ \frac{f_{x_t}(L_{t,i-1} \oplus \pi(x_t))}{\ell(\pi(x_t))}] \\
	& - \frac{ f_{x_t}(L_{t,i}) - f_{x_t}(L_{t,i-1}) }{\ell(s_{t,i})} )\\
	= & \mathbb{E}_{\pi \sim U(L^*_{\pi})}  \frac{1}{T}\sum_{t=1}^T\sum_{i=1}^{|L_t|} w_i (\frac{f_{x_t}(L_{t,i-1} \oplus \pi(x_t))}{\ell(\pi(x_t))} \\
		& - \frac{ f_{x_t}(L_{t,i}) - f_{x_t}(L_{t,i-1}) }{\ell(s_{t,i})} )\\
	\leq & \max_{\pi \in \Pi}  \frac{1}{T}\sum_{t=1}^T\sum_{i=1}^{|L_t|} w_i (\frac{f_{x_t}(L_{t,i-1} \oplus \pi(x_t))}{\ell(\pi(x_t))} \\
			& - \frac{ f_{x_t}(L_{t,i}) - f_{x_t}(L_{t,i-1}) }{\ell(s_{t,i})} )\\
	= & R/T\\
	\end{array}
	\end{displaymath}
	Hence combining with the previous result proves the theorem.
\end{proof}

%

\begin{corSCPDefKnapsack}
If we run an online learning algorithm on the sequence of convex losses $C_t$ instead of $\ell_t$, then after $T$ iterations, for any $\delta \in (0,1)$, we have that with probability at least $1-\delta$:
\begin{displaymath}
F(\overline{\pi}) \geq (1-1/e) F(\tilde{L}^*_{\pi}) - \frac{\tilde{R}}{T} - 2 \sqrt{\frac{2 \ln(1/\delta)}{T}} - \mathcal{G}
\end{displaymath}

where $\tilde{R}$ is the regret on the sequence of convex losses $C_t$, and $\mathcal{G} = \frac{1}{T}[\sum_{t=1}^T (\ell_t(\pi^t) - C_t(\pi^t)) + \min_{\pi \in \Pi} \sum_{t=1}^T C_t(\pi) - \min_{\pi' \in \Pi} \sum_{t=1}^T \ell_t(\pi')]$ is the  ``convex optimization gap'' that measures how close the surrogate losses $C_t$ are to minimizing the cost-sensitive losses $\ell_t$.
\end{corSCPDefKnapsack}

\begin{proof}
Follows immediately from Theorem \thmSCPKnapsack  using the definition of $R$, $\tilde{R}$ and $\mathcal{G}$, since $\mathcal{G} = \frac{R - \tilde{R}}{T}$
\end{proof}

\end{document}